\documentclass[letterpaper, 10 pt, conference]{ieeeconf}
\IEEEoverridecommandlockouts

\usepackage[colorlinks]{hyperref}
\hypersetup{
    colorlinks,
    linkcolor=black,
    citecolor=black,
    filecolor=magenta,
    urlcolor=cyan,
}

\usepackage{resizegather}

\usepackage{cite}
\usepackage{amsmath,amssymb,amsfonts,mathrsfs}

\DeclareMathOperator*{\argmin}{arg\,min}
\usepackage{algorithmic}
\usepackage{graphicx}
\usepackage{textcomp}
\usepackage{xcolor}
\usepackage{tcolorbox}

\usepackage{tablefootnote}
\usepackage{threeparttable}
\usepackage{caption, subcaption}

\usepackage{tikz}
\usepackage{graphicx}
\usepackage{tkz-euclide}

\usetikzlibrary{positioning}
\usetikzlibrary{shapes}
\usetikzlibrary{shapes.misc}
\usetikzlibrary{shapes.geometric}
\usetikzlibrary{plotmarks}
\usetikzlibrary{intersections}
\usetikzlibrary{calc}
\usetikzlibrary{fit}
\usetikzlibrary{patterns,tikzmark}
\usetikzlibrary{matrix,decorations.pathreplacing,calc}

\tikzset{cross/.style={cross out, draw, 
         minimum size=2*(#1-\pgflinewidth), 
         inner sep=0pt, outer sep=0pt}}

\newcommand{\state}[0]{x}
\newcommand{\prel}[0]{p_{\rm{rel}}}
\newcommand{\vrel}[0]{v_{\rm{rel}}}
\newcommand{\preldot}[0]{\dot{p}_{\rm{rel}}}
\newcommand{\vreldot}[0]{\dot{v}_{\rm{rel}}}

\newcommand{\norm}[1]{\left\lVert#1\right\rVert}

\newcommand{\vertiii}[1]{{\left\vert\kern-0.25ex\left\vert\kern-0.25ex\left\vert #1 
    \right\vert\kern-0.25ex\right\vert\kern-0.25ex\right\vert}}

\newtheorem{theorem}{Theorem}

\newtheorem{remark}{Remark}

\newtheorem{definition}{Definition}

\makeatletter
\renewcommand{\fps@figure}{htp}
\renewcommand{\fps@table}{htp}
\makeatother

\def\BibTeX{{\rm B\kern-.05em{\sc i\kern-.025em b}\kern-.08em
    T\kern-.1667em\lower.7ex\hbox{E}\kern-.125emX}}
    
\begin{document}

\title{Polygonal Cone Control Barrier Functions (PolyC2BF) for safe navigation in cluttered environments}

\author{Manan Tayal, Shishir Kolathaya
\thanks{Manan is supported by the Prime Minister Research Fellowship (PMRF).
The research was supported by the SERB grant CRG/2021/008115.
}
\thanks{$^{1}$Robert Bosch Center for Cyber Physical Systems (RBCCPS), Indian Institute of Science (IISc), Bengaluru.
{\tt\scriptsize \{manantayal, shishirk\}@iisc.ac.in}
}%
}

\maketitle
\begin{abstract}
In fields such as mining, search and rescue, and archaeological exploration, ensuring real-time, collision-free navigation of robots in confined, cluttered environments is imperative. Despite the value of established path planning algorithms, they often face challenges in convergence rates and handling dynamic infeasibilities. Alternative techniques like collision cones struggle to accurately represent complex obstacle geometries. This paper introduces a novel category of control barrier functions, known as Polygonal Cone Control Barrier Function (PolyC2BF), which addresses overestimation and computational complexity issues. The proposed PolyC2BF, formulated as a Quadratic Programming (QP) problem, proves effective in facilitating collision-free movement of multiple robots in complex environments. The efficacy of this approach is further demonstrated through PyBullet simulations on quadruped (unicycle model), and crazyflie 2.1 (quadrotor model) in cluttered environments. 
\end{abstract}


\section{Introduction}
\label{section: Introduction}
\par Ensuring real-time collision-free navigation of robots within confined, cluttered environments is of paramount importance in critical fields such as mining, search and rescue operations in structurally compromised buildings, and archaeological exploration of sites featuring narrow passageways or constrained spaces. It is imperative to uphold the highest standards of safety and task efficiency in these scenarios. While established optimal path planning algorithms like A*\cite{A-star}, RRT*\cite{RRT-star}, and PRM \cite{844107} offer valuable insights, they grapple with slow convergence rates, rendering them less effective in promptly identifying optimal paths, especially in intricate environments. Additionally, they do not address the dynamic infeasibility challenges encountered during the execution of paths generated by higher-level planners. Approaches like kinodynamic RRT* \cite{Kinodynamic-RRT} and LQR-RRT*\cite{LQR-RRT} play a pivotal role in determining appropriate steering inputs for seamless transitions between vertices within the sampling graph. Nevertheless, these techniques fall short in their ability to conduct dynamic collision assessments that accurately accommodate the nonlinear dynamics of the robot.

\begin{figure}[t]
    \centering
    
    \includegraphics[width=0.52\linewidth]{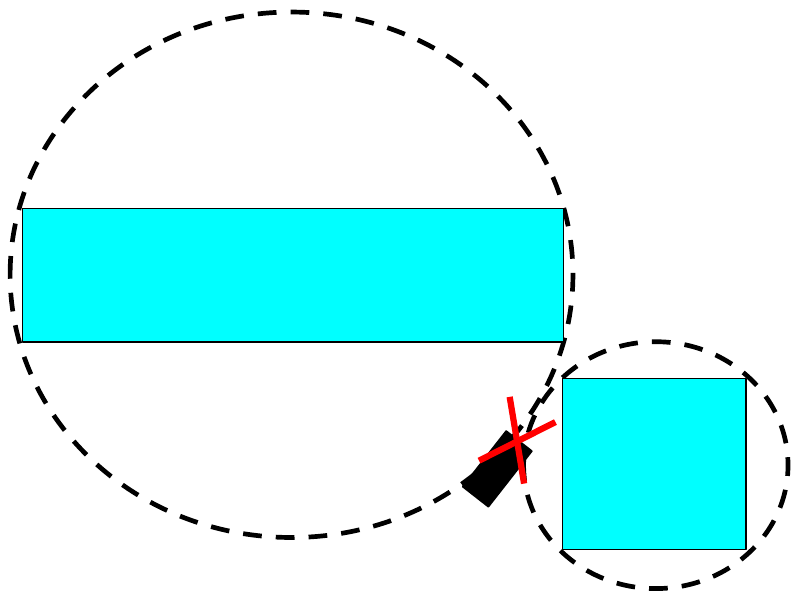}
    \includegraphics[width=0.44\linewidth]{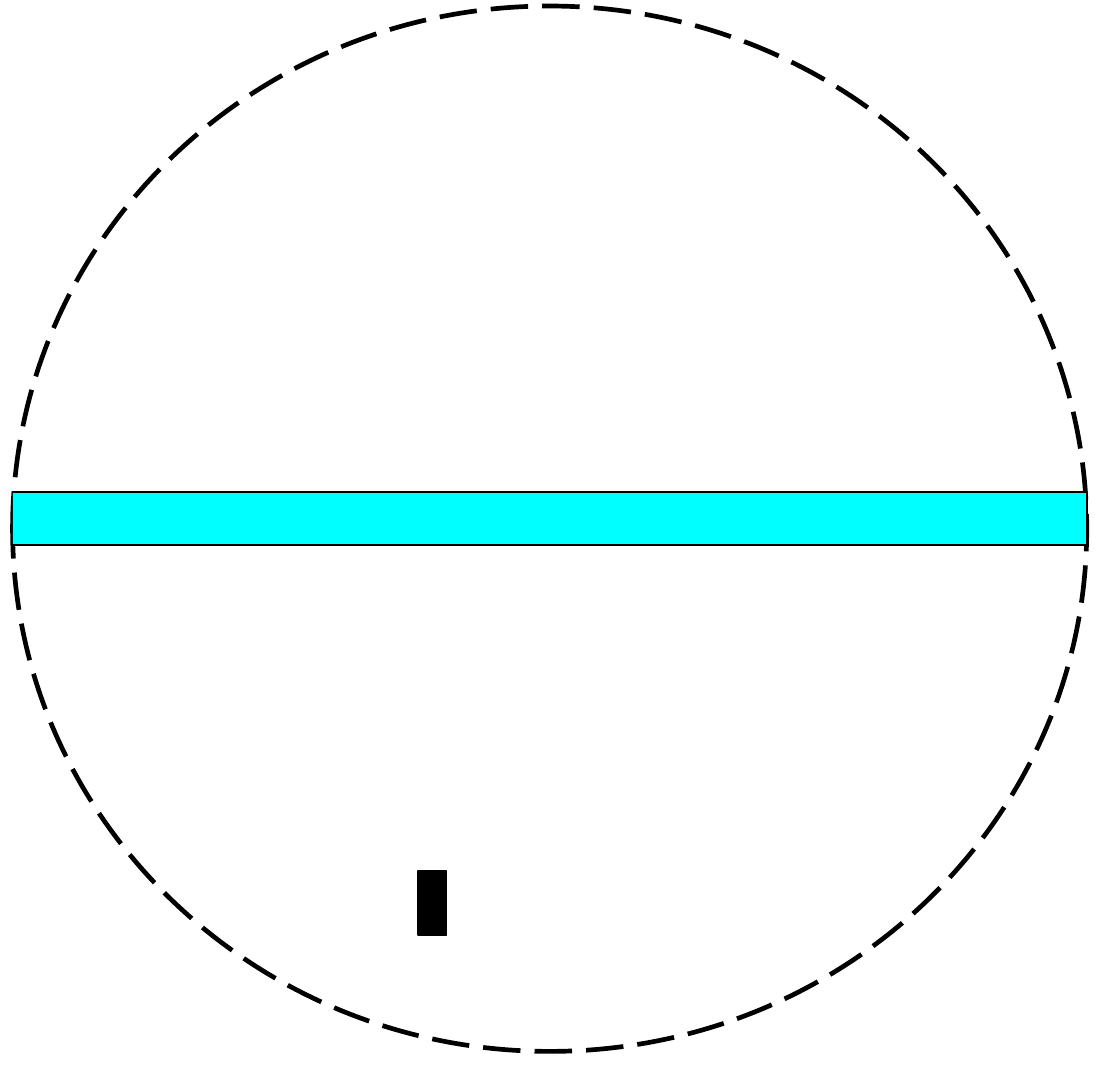}
\caption{Disadvantages of Collision Cones: Collision Avoidance in cluttered environment (left) and against a long wall (right) }
\label{fig: disadvantages of CC}
\end{figure}

Alternative collision avoidance techniques like collision cones \cite{Fiorini1993, doi:10.1177/027836499801700706, 709600} are not without notable limitations. Within cluttered environments, the assumption of circular obstacles can result in imprecise collision predictions and sub-optimal path choices due to this oversimplified representation of obstacles. This deficiency is even apparent in scenarios where the robot is endeavoring to evade collision with an extended wall, potentially leading to the incorrect perception that the robot is already within the obstacle, ultimately resulting in a failure state. This phenomenon is visually exemplified in Fig. \ref{fig: disadvantages of CC}. This drawback highlights the importance of more sophisticated collision avoidance methods that can accurately account for the complex geometry of obstacles in the environment.

The Collision Cone Control Barrier Functions \cite{C3BF-UGV, C3BF-UAV, C3BF-Legged}, an integration of Collision Cones and Control Barrier Functions (CBFs) \cite{7040372}\cite{Ames_2017} principles  also confront the challenges observed in collision cones. CBFs are instrumental in establishing formal safety guarantees, particularly in the context of robust collision avoidance with obstacles. These functions delineate a secure state set through inequality constraints and frame it as a quadratic programming (QP) problem to ensure the ongoing invariance of these sets over time.

A notable advantage of employing CBF based QP, when compared to other contemporary techniques like model predictive control (MPC) \cite{10178219}, reference governor \cite{6859176}, reachability analysis \cite{8263977}, and artificial potential fields \cite{12131}, lies in their efficacy in real-time applications within complex dynamic environments \cite{Singletary2021ComparativeAO, https://doi.org/10.48550/arxiv.2106.13176}. This translates to the ability to compute optimal control inputs at a notably high frequency using off-the-shelf electronics. Furthermore, CBFs can serve as a rapid safety filter when overlaid onto existing path planning controllers \cite{9682803}.

In scenarios requiring precise obstacle avoidance with minimal clearance, it is imperative to model obstacles as polyhedra. \cite{9812334, 9867246} This work accomplishes real-time collision avoidance by leveraging convex and non-convex optimizations in conjunction with control barrier functions. Nonetheless, it is worth noting that optimization-based methodologies typically encounter a significant rise in computational complexity as the number of robots involved increases. This scalability challenge can prove demanding when implementing such approaches in real-time for distributed multi-robot systems. 

Hence, this paper presents a novel category of control barrier functions referred to as Polygonal Cone Control Barrier Function (PolyC2BF). This advancement builds upon the existing Collision Cone Control Barrier Function (C3BF) and rectifies the issue of overestimation of obstacle and robot dimensions when represented as circles. Additionally, the proposed PolyC2BF mitigates the computational complexity associated with optimization-based approaches. It is demonstrated to facilitate collision-free movement of multiple robots in complex, confined environments.

Main contributions of our work are as follows:

\begin{itemize}
    \item We formulate a direct method for safe trajectory tracking of non-holonomic ground and aerial vehicles based on polygonal cone control barrier functions expressed through a quadratic program to traverse in cluttered environment and tight spaces.
    \item We verify the polygonal cone CBF (PolyC2BF) based safety filter in simulations and compare it with the collision cone CBF (C3BF), to show how the former is better in terms of safe navigation in cluttered environment. 
\end{itemize}


\section{Background}
\label{section: Background}
In this section, we will introduce the necessary definitions and the relevant background for Control Barrier Functions (CBFs), the collision cone control barrier function and CBF-based safety filters.

Consider a nonlinear control system in affine form:
\begin{equation}
	\dot{\state} = f(\state) + g(\state)u
	\label{eqn: affine control system}
\end{equation}
where $\state \in \mathcal{D} \subseteq \mathbb{R}^n$ is the state of system, and $u \in \mathbb{U} \subseteq \mathbb{R}^m$ the input for the system. Assume that the functions $f: \mathbb{R}^n \rightarrow \mathbb{R}^n$ and $g: \mathbb{R}^n \rightarrow \mathbb{R}^{n \times m}$ are locally Lipschitz continuous functions. 

The safe set $\mathcal{C}$ is defined as the \textit{super-level set} of a continuously differentiable function $h:\mathcal{D}\subseteq \mathbb{R}^n \rightarrow \mathbb{R}: \mathcal{C} = \{ \state \in \mathcal{D} : h(\state) \geq 0\}$ where the boundary of the safe set is $\partial\mathcal{C} = \{ \state \in \mathcal{D} : h(\state) = 0\}$ with $\frac{\partial h}{\partial \state}(\state) \neq 0\; \forall \state \in \partial \mathcal{C}$ and the interior of the safe set is $\text{Int}\left(\mathcal{C}\right) = \{ \state \in \mathcal{D} \subset \mathbb{R}^n : h(\state) > 0\}$. It is assumed that $\text{Int}\left(\mathcal{C}\right)$ is non-empty and $\mathcal{C}$ has no isolated points, i.e. $\text{Int}\left(\mathcal{C}\right) \neq \phi$ and $\overline{\text{Int}\left(\mathcal{C}\right)} = \mathcal{C}$. The goal is to minimally modify a given control policy $\pi(\state, t)$ with a safety based on a continuous time-CBF filter such that the system is safe (i.e. the system's state $\state$ stays in safe set $\mathcal{C}$ if it starts inside $\mathcal{C}$).

\begin{definition}[Extended class-$\mathcal{K}$ functions]{\it
\label{definition: extended class k definition}
A continuous function $\kappa : [0, d) \rightarrow [0, \infty)$ for some $d > 0$ is said to belong to class-$\mathcal{K}$ if it is strictly increasing and $\kappa(0) = 0$. Here, $d$ is allowed to be $\infty$. The same function can be extended to the interval $\kappa: (-b,d)\to (-\infty, \infty)$ with $b>0$ (which is also allowed to be $\infty$), in which case we call it the extended class $\mathcal{K}$ function.}
\end{definition}

\begin{definition}[Control barrier function (CBF)]{\it
\label{definition: CBF definition}
Given the set $\mathcal{C} \subset \mathcal{D}$ be the superlevel set of a continuously differentiable function $h:\mathcal{D}\subseteq \mathbb{R}^n \rightarrow \mathbb{R}$ with $\frac{\partial h}{\partial \state}(\state) \neq 0\; \forall \state \in \partial \mathcal{C}$, the function $h$ is called the control barrier function (CBF) defined on the set $\mathcal{D}$, if there exists an extended \textit{class} $\mathcal{K}$ function $\kappa$ such that for all $\state \in \mathcal{D}$:

\begin{equation}
\begin{aligned}
    \underbrace{\text{sup}}_{ u \in \mathbb{U}}\! \left[\underbrace{\mathcal{L}_{f} h(\state) + \mathcal{L}_g h(\state)u} \iffalse+ \frac{\partial h}{\partial t}\fi_{\dot{h}\left(\state, u\right)} \! + \kappa\left(h(\state)\right)\right] \! \geq \! 0
\end{aligned}\label{eqn:CBF_conditions}
\end{equation}
where $\mathcal{L}_{f} h(\state) = \frac{\partial h}{\partial \state}f(\state)$ and $\mathcal{L}_{g} h(\state)= \frac{\partial h}{\partial \state}g(\state)$ are the Lie derivatives. 
}
\end{definition}

Given this definition of a CBF, we know from \cite{Ames_2017} and \cite{8796030} that any Lipschitz continuous control law $k(\state)$ satisfying the inequality: $\dot{h} + \kappa( h )\geq 0$ ensures safety of $\mathcal{C}$ if $x(0)\in \mathcal{C}$, and asymptotic convergence to $\mathcal{C}$ if $x(0)$ is outside of $\mathcal{C}$. 

\begin{definition}[Collision Cone CBFs \cite{C3BF-UGV,C3BF-UAV} ]{\it
\label{definition: c3bf definition}
This approach combines the idea of potential unsafe directions given by collision cone as unsafe set to formulate a CBF.
Consider the following CBF candidate:
\begin{equation}
    h(\state, t) = < \prel, \vrel> + \| \prel\|\| \vrel\|\cos\phi ,
    \label{eqn:CC-CBF}
\end{equation}
where $\prel$ is the relative position vector between the body center of the ego vehicle and the center of the obstacle, $\vrel$ is the relative velocity, $<\cdot , \cdot>$ is the dot product of 2 vectors and $\phi$ is the half angle of the cone, the expression of $\cos\phi$ is given by $\frac{\sqrt{\|\prel\|^2 - r^2}}{\|\prel\|}$. The proposed constraint simply ensures that the angle between $\prel, \vrel$ is less than $180^\circ - \phi$.
}
\end{definition}
Precise mathematical definitions for $\prel, \vrel$ will be given in the next section.

Using the CBF conditions, we can define an input set:
\begin{equation}
\begin{aligned}
\label{eqn: U_CBF}
\mathbb{U}_{cbf}(\state) = \{ u \in \mathbb{U} \subseteq \mathbb{R}^m :\dot{h} + \kappa\left(h(\state)\right) \! \geq \! 0 \}\\
\end{aligned}
\end{equation}
that renders the system safe. This requires the selection of an extended class-$\mathcal{K}$ function $\kappa$ that yields a non-empty $\mathbb{U}_{cbf}(\state)$ for all $\state \in \mathcal{C}$.
For a given control policy $\pi(\state,t)$ that may not be safe, one can formulate a quadratic program (QP) which satisfy the CBF condition \eqref{eqn:CBF_conditions} to modify this input in a minimal way:

\begin{equation}
\begin{aligned}
\label{eqn: CBF QP}
u^{*}(x,t) &= \argmin_{u \in \mathbb{U} \subseteq \mathbb{R}^m} \norm{u - \pi(x,t)}^2\\
\quad & \textrm{s.t. } \mathcal{L}_f h(x) + \mathcal{L}_g h(x)u + \kappa \left(h(x)\right) \geq 0\\
\end{aligned}
\end{equation}
This is called the Control Barrier Function based Quadratic Program (CBF-QP). If $\mathbb{U}=\mathbb{R}^m$, then the QP is feasible, and the explicit solution is given by
\begin{equation*}
	u^{*}(x, t) = \pi(x, t) + u_{safe}(x,t)
\end{equation*}
where $u_{safe}(x,t)$ is given by
\begin{multline}\label{eq:CBF-QP}
u_{safe}(x, t) \!=\!
	\begin{cases}
		0 & \text{for } \psi(x, t) \geq 0 \\
		-\frac{\mathcal{L}_{g}h(x)^T \psi(x, t)}{\mathcal{L}_{g}h(x)\mathcal{L}_{g}h(x)^T} & \text{for } \psi(x, t) < 0
	\end{cases}
\end{multline}

where $\psi (x,t) := \dot{h}\left(x, \pi(x, t)\right) + \kappa \left(h(x)\right)$. The sign change of $\psi$ yields a switching type of a control law.

\section{Polygonal Cone Control Barrier Functions (PolyC2BF)}
\label{section: Collision Cone CBF}
\begin{figure}
    \centering
    \includegraphics[width=0.47\linewidth]{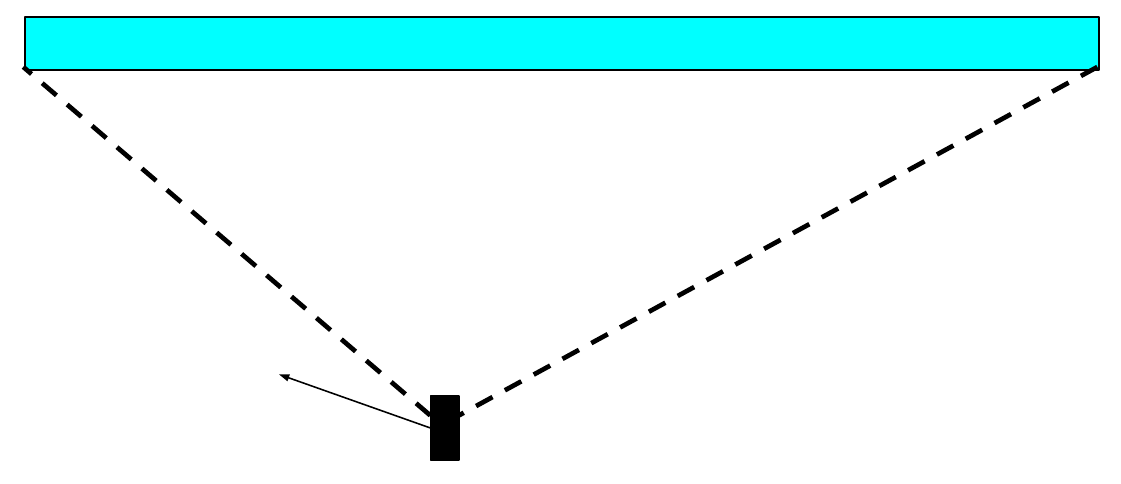}
    \includegraphics[width=0.49\linewidth]{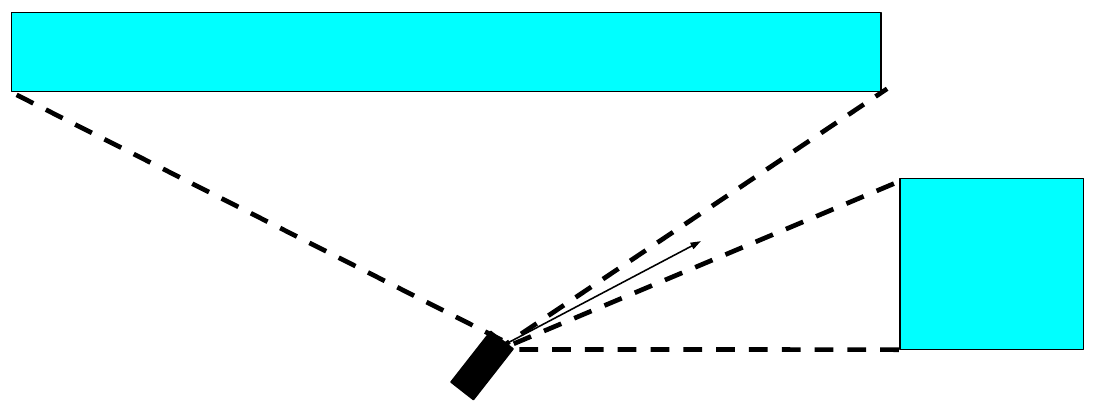}
    \includegraphics[width=0.47\linewidth]{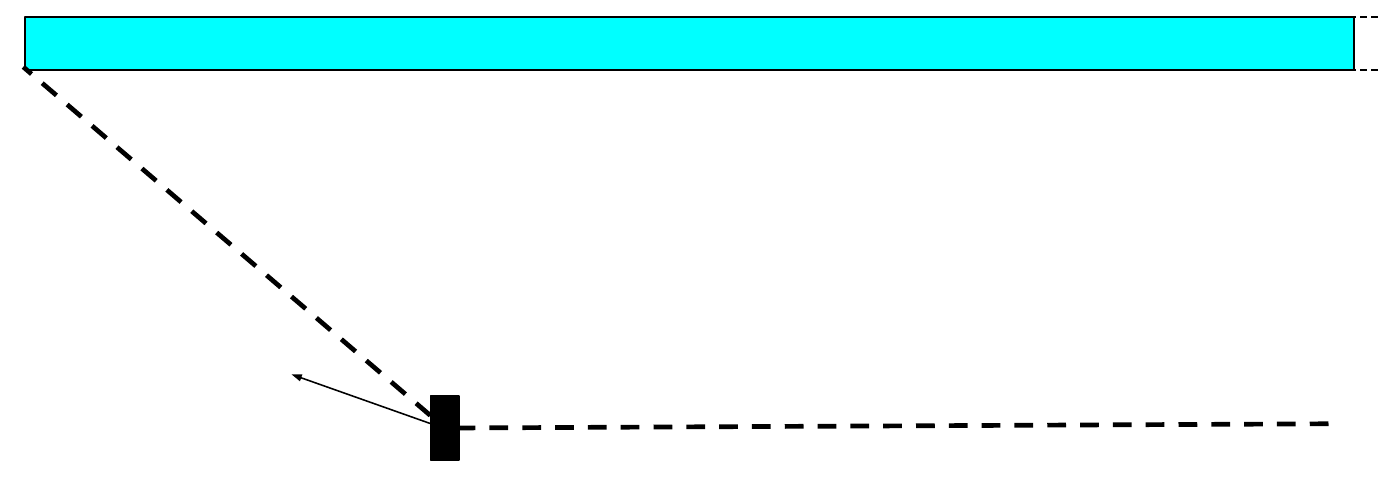}
\caption{Polygonal Collision Cone using vertices of the polygonal obstacles}
\label{fig: Solution}
\end{figure}

Assuming the polygonal obstacles as circle/ellipses is not a great idea as shown in Fig \ref{fig: disadvantages of CC}, especially in cases of very long walls or cluttered environment. However, instead if we consider the vertices/edges of the polygon to construct a cone to be avoided, we may be able to resolve the issue. Fig \ref{fig: Solution} shows the construction of Polygonal Collision Cones for long wall, semi-infinite wall and cluttered environments.  

\subsection{Polygonal Cone CBF (PolyC2BF)}
Given the outlined limitations associated with assuming obstacles as circles or ellipses, particularly for polygonal obstacles, and the methodology for constructing a cone around a polygonal obstacle, we propose integrating these concepts with Control Barrier Functions to formulate a real-time safe controller tailored for these scenarios.

\begin{figure}
    \centering
    \includegraphics[width=0.9\linewidth]{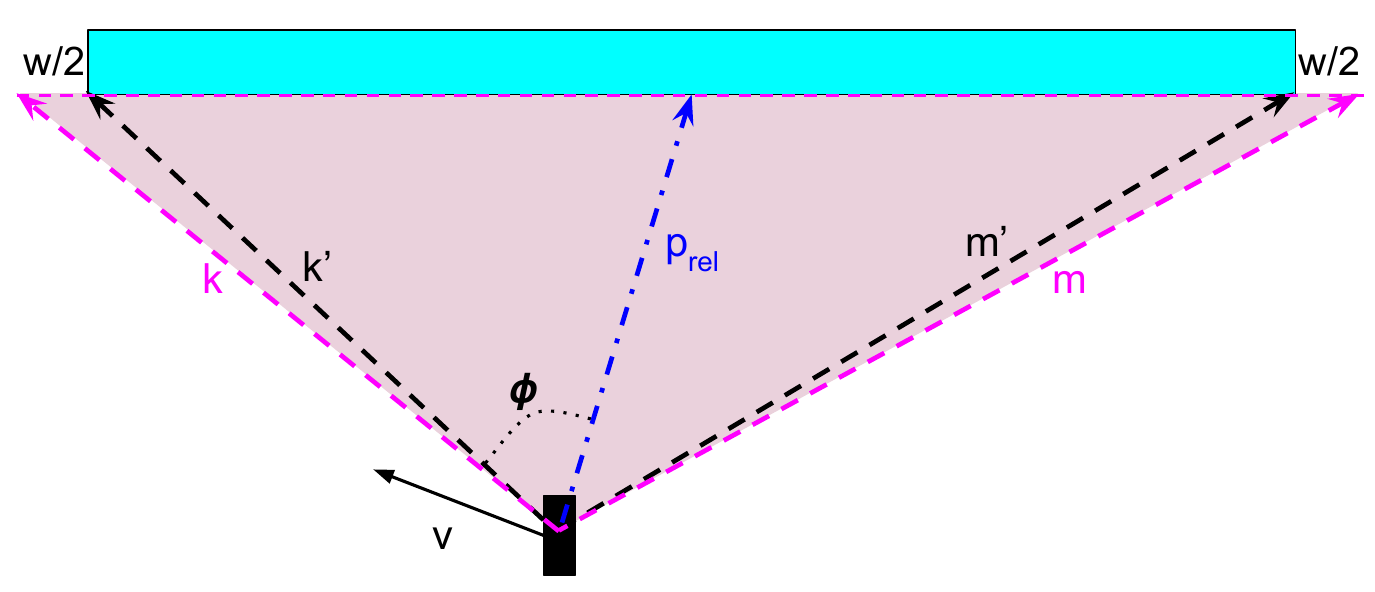}
    \includegraphics[width=0.8\linewidth]{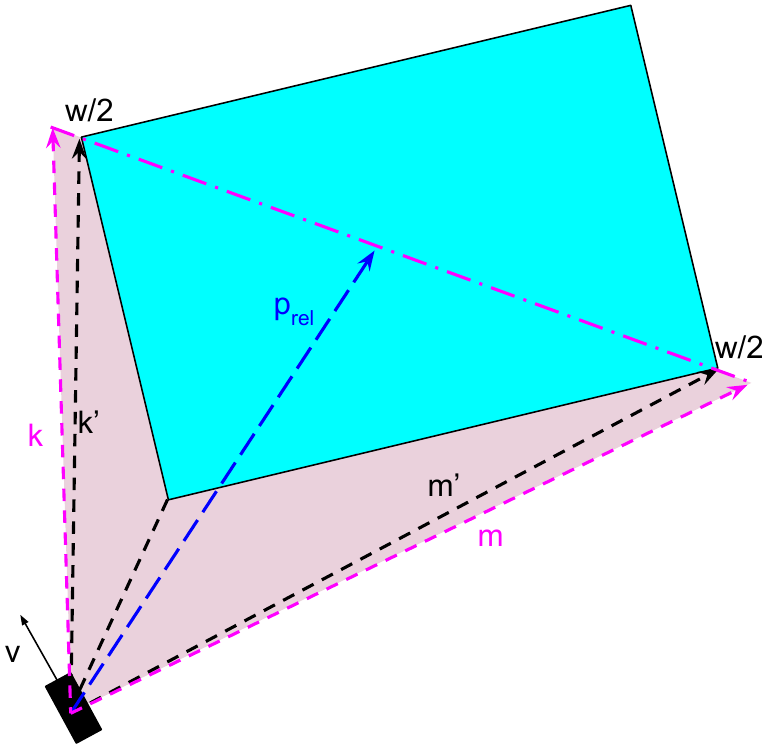}
    \caption{Construction of Polygonal Cone to avoid polygonal obstacle}
    \label{fig:Safe Unsafe Sets Ellipse}
\end{figure}

For polygonal obstacles, the cone can be directly constructed using their vertices. These vertices are selected to ensure comprehensive coverage of the entire polygon. The relative position vector ($\prel$) is oriented towards the midpoint of the line connecting these two vertices. Let's denote the vector that forms the most substantial angle with $\prel$ as $k'$, and the other one as $m'$. To account for the width/shape of the ego vehicle ($w$) \footnote{$w$ can be predetermined (e.g. the diagonal of the vehicle) or can be dynamically adjusted (e.g. $w$ may change as a function of the orientation of the ego-vehicle).}, the line joining these two vertices is extended by $\frac{w}{2}$ on both sides, forming vectors $k$ and $m$ (corresponding to vectors $k'$ and $m'$, respectively) as depicted in Fig \ref{fig:Safe Unsafe Sets Ellipse}.

\begin{remark}
    \textit{The vector connecting the vertex which is closer to the ego vehicle, forms the bigger angle with $\prel$. Thus, in the analysis shown next, we will try to avoid a cone with half angle $\phi$, is the angle between vectors $k$ (vector connecting the closer vertex) and $\prel$.}
\end{remark}

This novel approach of avoiding the pink cone region using Control Barrier Functions, gives rise to \textbf{Polygonal Cone Control Barrier Functions (PolyC2BFs)}. The proposed CBF candidate is as follows:
%
\begin{equation}
    h(\state, t) = <\prel,\vrel> + \|\prel\| \|\vrel\| \cos\phi
    \label{eqn:PolyC3BF}
\end{equation}
where, the expression of $\cos\phi$ is given by $\frac{<\prel,k> } {\|\prel\| \|k\|}$.

We now consider an acceleration controlled non-holonomic ground vehicle (unicycle model) described by
\begin{equation}
	\underbrace{\begin{bmatrix}
		\dot{x}_p \\
		\dot{y}_p \\
		\dot{\theta} \\
		\dot{v} \\
            \dot{\omega}
	\end{bmatrix}}_{\dot{\state}}
	=
        \underbrace{\begin{bmatrix}
            v\cos\theta\\
            v\sin\theta\\
            \omega \\
            0 \\
            0
        \end{bmatrix}}_{f(\state)}
	+
	\underbrace{\begin{bmatrix}
            0 & 0 \\
            0 & 0 \\
            0 & 0 \\
            1 & 0 \\
            0 & 1
	\end{bmatrix}}_{g(\state)}
	\underbrace{\begin{bmatrix}
		a \\
		\alpha
	\end{bmatrix}}_{u}
    \label{eqn:Acceleration controlled vehicle}
\end{equation}

where state variables denoted as $x_p$, $y_p$, $\theta$, $v$, and $\omega$, representing pose, linear velocity, and angular velocity, respectively. Linear acceleration $(a)$ and angular acceleration $(\alpha)$ serve as the control inputs.

We first obtain the relative position vector between the body center of the vehicle and the center of the obstacle. 
Therefore, we have
\begin{align}\label{eq:positionvectorunicycle}
    \prel := \begin{bmatrix}
        c_x - (x_p + l \cos(\theta)) \\
        c_y - (y_p + l \sin(\theta))
    \end{bmatrix}
\end{align}
Here $l$ is the distance of the body center from the differential drive axis. $c_x,c_y$ represents the center of obstacle as a function of time. Also, since the obstacles are of constant velocity, we have $\Ddot{c}_x= \Ddot{c}_y= 0$. We obtain its velocity as
\begin{align}\label{eq:velocityvectorunicycle}
    \vrel := \begin{bmatrix}
        \dot c_x - (v \cos (\theta) - l \sin(\theta)*\omega) \\
        \dot c_y - (v \sin (\theta) + l \cos(\theta)*\omega)
    \end{bmatrix}.
\end{align}

We have the following first result of the paper:
\begin{theorem}\label{thm:unicycletheorem}{\it
The proposed CBF candidate \eqref{eqn:PolyC3BF} with $\prel,\vrel$ defined by \eqref{eq:positionvectorunicycle}, \eqref{eq:velocityvectorunicycle} is a valid CBF defined for the set $\mathcal{D}$.}
\end{theorem}
\begin{proof}
Taking the derivative of \eqref{eqn:PolyC3BF} yields
\begin{align}
\dot h = &  < \preldot, \vrel > + < \prel, \vreldot >  \nonumber \\
 & + < \vrel, \vreldot > \frac{<\prel,k>}{\|\vrel\| \|k\|} \nonumber \\
 & + < \preldot, \hat{k} > \|\vrel\| 
 + < \prel, \dot{\hat{k}} > \|\vrel\| .
 \label{eqn:h_derivative}
\end{align}
where, $\hat{k}$ means $k/ \|k\|$.
Further $\preldot  = \vrel$ and
\begin{align*}
    \vreldot = \begin{bmatrix}
        - a \cos \theta + v (\sin \theta) \omega + l (\cos \theta) \omega^2 + l (\sin \theta) \alpha \\
        -a \sin \theta - v (\cos \theta) \omega + l (\sin \theta) \omega^2 - l (\cos \theta) \alpha
    \end{bmatrix}. \nonumber
\end{align*}
Given $\vreldot$ and $\dot h$, we have the following expression for $\mathcal{L}_g h$:
\begin{align}
    \mathcal{L}_g h = \begin{bmatrix}
        < \prel + \vrel \frac{< \prel, \hat{k} >}{\|\vrel\|}, \begin{bmatrix}
            - \cos \theta \\
            - \sin \theta
        \end{bmatrix}>  \\
                < \prel + \vrel \frac{< \prel, \hat{k} >}{\|\vrel\|}, \begin{bmatrix}
            l \sin \theta \\
            - l \cos \theta
        \end{bmatrix}> 
    \end{bmatrix}^T,
\end{align}
By employing analogous reasoning as demonstrated in the proof of \cite[Theorem 1]{C3BF-UGV}, it can be ascertained that $\mathcal{L}_gh$ is always a non-zero matrix. Thus, in accordance with \cite[Theorem 8]{XU201554}, we can deduce that the resultant Quadratic Programming problem (QP), as defined in \eqref{eq:CBF-QP}, exhibits Lipschitz continuity. Consequently, we can formulate CBF-QPs utilizing the proposed CBF, as expressed in \eqref{eqn:PolyC3BF}, to ensure collision avoidance.
\end{proof}

\begin{figure*}
    \centering
    \includegraphics[width=0.35\linewidth]{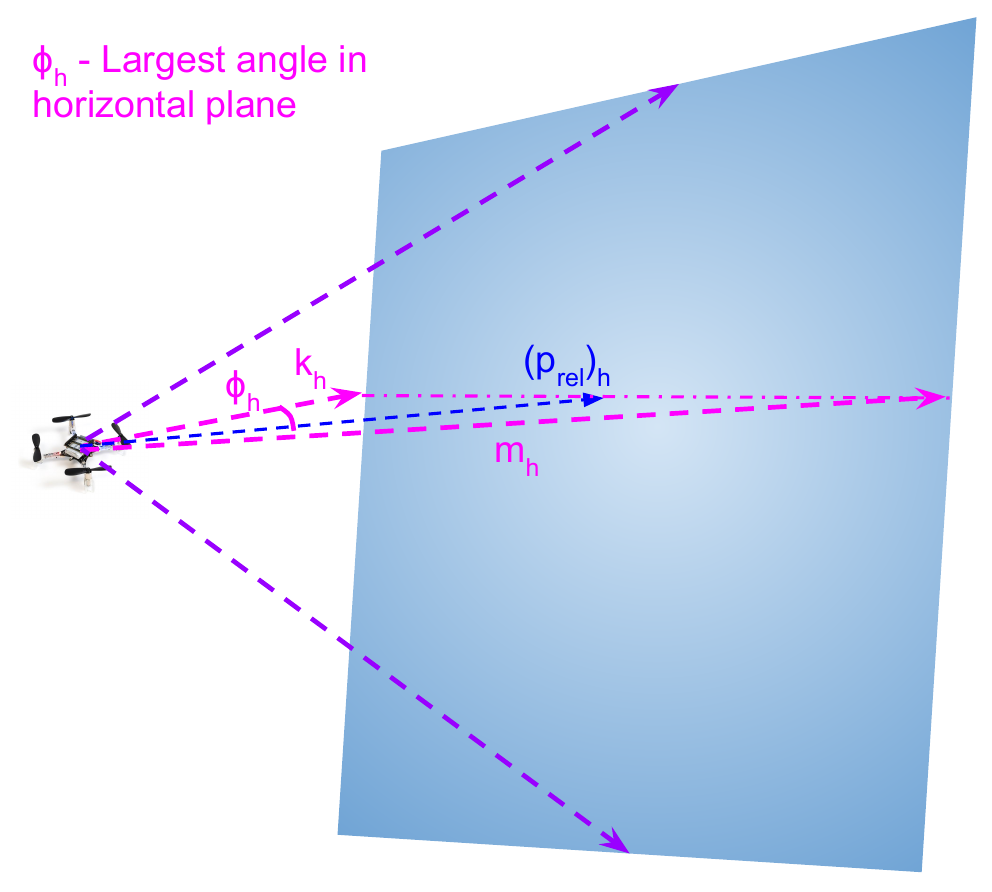}
    \includegraphics[width=0.35\linewidth]{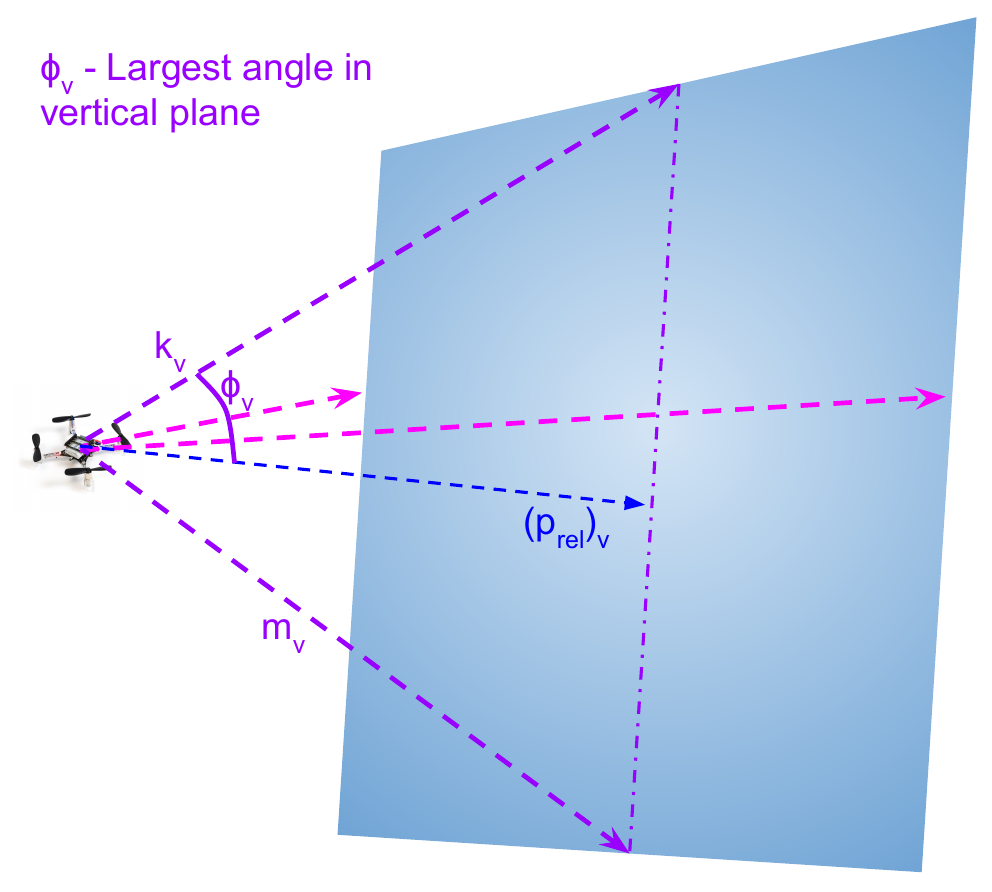}
    \caption{Construction of Polygonal Cone for an Aerial Vehicle}
    \label{fig: 3d-case}
\end{figure*}

Next, let's consider the aerial vehicle (quadrotor model). 
In this case we will have to consider construction of polygonal cones in both horizontal and vertical planes thus forming vectors $k_{h}$ and $k_v$ for horizontal and vertical planes respectively, as shown in Fig \ref{fig: 3d-case}. In case, $\|k_h \| < \|k_v \|$ , we will consider $k = k_h$ and $\prel = (\prel)_h$, similarly, if $\|k_h \| > \|k_v \|$ , we will consider $k = k_v$ and $\prel = (\prel)_v$. This will be similar to the projection case in \cite{C3BF-UAV}.

A quadrotor has four propellers, which provides upward thrusts of $(f_1, f_2, f_3, f_4)$ and act as inputs. The states needed to describe the quadrotor system is given by $x = [x_p, y_p, z_p, \dot{x}_p, \dot{y}_p, \dot{z}_p,  \phi, \theta, \psi, {\omega}_{1}, {\omega}_{2}, {\omega}_{3}]$ . The quadrotor dynamics is as follows:
\begin{equation}
\label{eqn:quadrotor_model}
	\underbrace{\begin{bmatrix}
		\dot{x}_p \\
		\dot{y}_p \\
            \dot{z}_p \\
            \ddot{x}_p \\
		\ddot{y}_p \\
            \ddot{z}_p \\
		\dot{\phi} \\
            \dot{\theta} \\
            \dot{\psi} \\
		\dot{\omega}_{1} \\
            \dot{\omega}_{2} \\
            \dot{\omega}_{3}
	\end{bmatrix}}_{\dot{\state}}
	=
	\underbrace{\begin{bmatrix}
		\dot{x}_p \\
		\dot{y}_p \\
            \dot{z}_p \\
            0 \\
            0 \\
            -g \\
            \textbf{W}^{-1}
            \begin{bmatrix}
                \omega_{1} \\
                \omega_{2} \\
                \omega_{3}
            \end{bmatrix}\\
            \\
		  -I^{-1} \vec{\omega} \times I \vec{\omega}  \\
            .
	\end{bmatrix}}_{f(\state)}
	+
	\underbrace{\begin{bmatrix}
		\\
            \begin{bmatrix}
                0 & 0 & 0 & 0 \\
                0 & 0 & 0 & 0 \\
                0 & 0 & 0 & 0 
            \end{bmatrix}\\
            \\
            \frac{1}{m_Q}\textbf{R}
            \begin{bmatrix}
                0 & 0 & 0 & 0 \\
                0 & 0 & 0 & 0 \\
                1 & 1 & 1 & 1 
            \end{bmatrix}\\
            \\
            \begin{bmatrix}
                0 & 0 & 0 & 0 \\
                0 & 0 & 0 & 0 \\
                0 & 0 & 0 & 0 
            \end{bmatrix}\\
            \\
		I^{-1}L
            \begin{bmatrix}
                1 & 0 & -1 & 0 \\
                0 & 1 & 0 & -1 \\
                c_{\tau} & -c_{\tau} & c_{\tau} & -c_{\tau} 
            \end{bmatrix}
	\end{bmatrix}}_{g(\state)}
	\underbrace{\begin{bmatrix}
		f_{1} \\
		f_{2} \\
            f_{3} \\
            f_{4} 
	\end{bmatrix}}_{u}
\end{equation}
$x_p$, $y_p$ and $z_p$ denote the coordinates of the vehicle’s centre of the base of the quadrotor in an inertial frame. $\phi$, $\theta$ and $\psi$ represents the (roll, pitch \& yaw) orientation of the quadrotor. \textbf{R} is the rotation matrix (from the body frame to the inertial frame), $m_Q$ is the mass of the quadrotor, $\textbf{W}$ is the transformation matrix for angular velocities from the inertial frame to the body frame, I is the inertia matrix and L is the diagonal length of quadrotor. $c_{\tau}$ is the constant
that determines the torque produced by each propeller. 

We have to obtain the relative position vector between the body center of the quadrotor and the intersection of the axis of the obstacle and the projection plane. Therefore, we have

\begin{align}\label{eqn:pos-vec-proj}
    \prel := \mathcal{P}\left (\begin{bmatrix}
        c_x \\
        c_y \\
        c_z
    \end{bmatrix}
    - \left (
    \begin{bmatrix}
        x_p \\
        y_p \\
        z_p
    \end{bmatrix}
    + \textbf{R} \begin{bmatrix}
        0 \\
        0 \\
        l
    \end{bmatrix}
    \right ) \right ).
\end{align}
Here $l$ is the distance of the body center from the base. $\mathcal{P}: \mathbb{R}^3 \to \mathbb{R}^3 $ is the projection operator, which can be assumed to be a constant\footnote{Note that the obstacles are always translating and not rotating. In addition, it is not restrictive to assume that the translation direction is always perpendicular to the cylinder axis. This makes the projection operator a constant.}.  $c_x,c_y,c_z$ represents the center of obstacle as a function of time. Also, since the obstacles are of constant velocity, we have $\Ddot{c}_x= \Ddot{c}_y= \Ddot{c}_z = 0$. We obtain its relative velocity as
\begin{align}\label{eqn:vel-vec-proj}
    \vrel := \frac{d({p}_{rel})}{dt} = \preldot
\end{align}

Now, we calculate the $\vreldot$ term which contains our inputs i.e. $(f_1, f_2, f_3, f_4)$, as follows:
\begin{equation}\label{eqn:vrel_dot}
    \frac{d}{dt}(\vrel) =
    \mathcal{P}(
            -  
            \textbf{R}
            \begin{bmatrix}
                0 & \frac{Ll}{I_{yy}} & 0 & \frac{-Ll}{I_{yy}} \\
                \frac{-Ll}{I_{xx}} & 0 & \frac{Ll}{I_{xx}} & 0 \\
                \frac{1}{m_Q} & \frac{1}{m_Q} & \frac{1}{m_Q} & \frac{1}{m_Q} 
            \end{bmatrix}
            \begin{bmatrix}
    		f_{1} \\
    		f_{2} \\
                f_{3} \\
                f_{4} 
    	\end{bmatrix} \\
        + \rm{additional} \: \rm{terms}). \nonumber
\end{equation}
Please note that the $\rm{additional} \: \rm{terms}$ in the above equation, refer to those terms that do not contain the input terms ($f_1,f_2,f_3,f_4$) and thus do not contribute to the calculation of $\mathcal{L}_g h$.
We have the following result with properties of $\prel$ and $\vrel$ similar to those in the projection case of \cite{C3BF-UAV}.

\begin{theorem}\label{thm:CC-CBF-3D}{\it
Given the quadrotor model \eqref{eqn:quadrotor_model}, the proposed CBF candidate \eqref{eqn:PolyC3BF} with $\prel,\vrel$ defined by \eqref{eqn:pos-vec-proj}, \eqref{eqn:vel-vec-proj} is a valid CBF defined for the set $\mathcal{D}$.}
\end{theorem}
\begin{proof}
%
The derivative of \eqref{eqn:PolyC3BF} is given by \eqref{eqn:h_derivative}.

By substituting for $\vreldot$ \eqref{eqn:vrel_dot} (only the term which contains the input) in $\dot h$ \eqref{eqn:h_derivative}, we have the following expression for $\mathcal{L}_g h$:
\begin{align}
    \mathcal{L}_g h = \begin{bmatrix}
        < \prel + \vrel \frac{< \prel, \hat{k} >}{\|\vrel\|}, 
            \textbf{R}\begin{bmatrix}
                0  \\
                \frac{-Ll}{I_{xx}}\\
                \frac{1}{m_Q}
            \end{bmatrix}>\\
                < \prel + \vrel \frac{< \prel, \hat{k} >}{\|\vrel\|}, 
            \textbf{R}\begin{bmatrix}
                \frac{Ll}{I_{yy}} \\
                0\\
                \frac{1}{m_Q}
            \end{bmatrix}>\\
         < \prel + \vrel \frac{< \prel, \hat{k} >}{\|\vrel\|}, 
            \textbf{R}\begin{bmatrix}
                0  \\
                \frac{Ll}{I_{xx}}\\
                \frac{1}{m_Q}
            \end{bmatrix}>\\
         < \prel + \vrel \frac{< \prel, \hat{k} >}{\|\vrel\|}, 
            \textbf{R}\begin{bmatrix}
                \frac{-Ll}{I_{yy}} \\
                0 \\
                \frac{1}{m_Q}
            \end{bmatrix}>
    \end{bmatrix}^T,
\end{align}

By employing analogous reasoning as demonstrated in the proof of \cite[Theorem 1]{C3BF-UAV}, it can be ascertained that $\mathcal{L}_gh$ is always a non-zero matrix, implying that it is a valid CBF. Consequently, QP formulation of PolyC2BF as expressed in \eqref{eqn:PolyC3BF}, ensures collision avoidance.
\end{proof}

Consider the point mass model, characterized as follows:
\begin{equation}
	\begin{bmatrix}
		\dot{p} \\
		\dot{v}
	\end{bmatrix}
	=
        \begin{bmatrix}
            0_{2x2} & I_{2x2} \\
            0_{2x2} & 0_{2x2}
        \end{bmatrix}
        \begin{bmatrix}
            p\\
            v\\
        \end{bmatrix}
	+
	\begin{bmatrix}
            0_{2x2} \\
            I_{2x2} 
	\end{bmatrix}
		u,
    \label{eqn:Acceleration controlled Point mass model}
\end{equation}
where $p$ = [$x_p$, $y_p$]$^T$, $v$ = [$v_x$, $v_y$]$^T$, and $u$ = [$a_x$, $a_y$]$^T$ $\in\mathbb{R}^2$ represent the position, velocity, and acceleration inputs, respectively. Analogous to Theorem \ref{thm:unicycletheorem}, similar results can be derived for this simplified point mass model. It is worth noting that the proposed PolyC2BF-QP stands as a valid Control Barrier Function (CBF), and its proof is straightforward.

\section{Simulations}
\label{section: Results and Discussions}
\par PolyC2BF-QP based controller has been validated on both quadruped (which can be assumed as an accelerated non holonomic vehicle (unicycle model) as in \cite{C3BF-Legged, molnar2021model}) and  quadrotor and the simulations were conducted on PyBullet \cite{coumans2019}, a python-based physics simulation engine. The Simulation runs on a computer with Ubuntu 22.04 with Intel i7-12800 HX CPU, with 16GB RAM and Nvidia 3070Ti GPU.


\begin{figure}
        \begin{subfigure}{0.9\linewidth}
        \includegraphics[width=\linewidth]{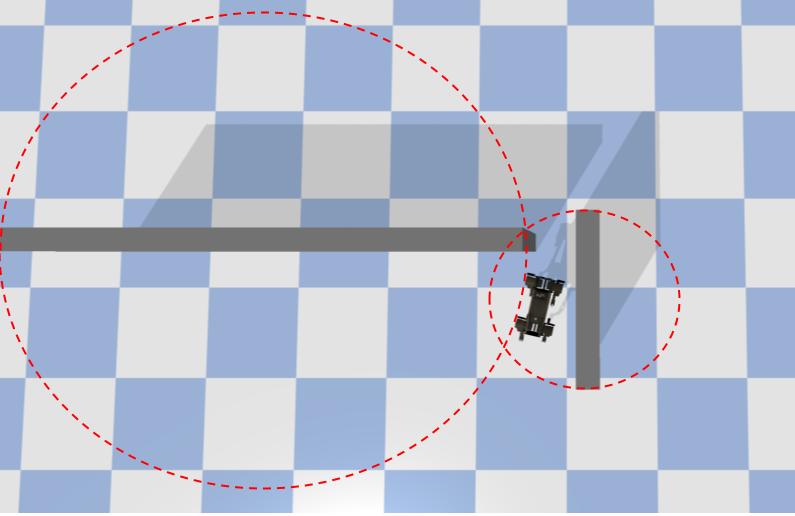}
        \end{subfigure}

\caption{Navigation in a cluttered environment using PolyC2BF in case of Quadruped}
\label{fig:cluttered_walk_quadruped}
\end{figure}

\begin{figure}
        \begin{center}
        \begin{minipage}{0.4\textwidth} 
        \begin{subfigure}{\linewidth}
        \includegraphics[width=\linewidth]{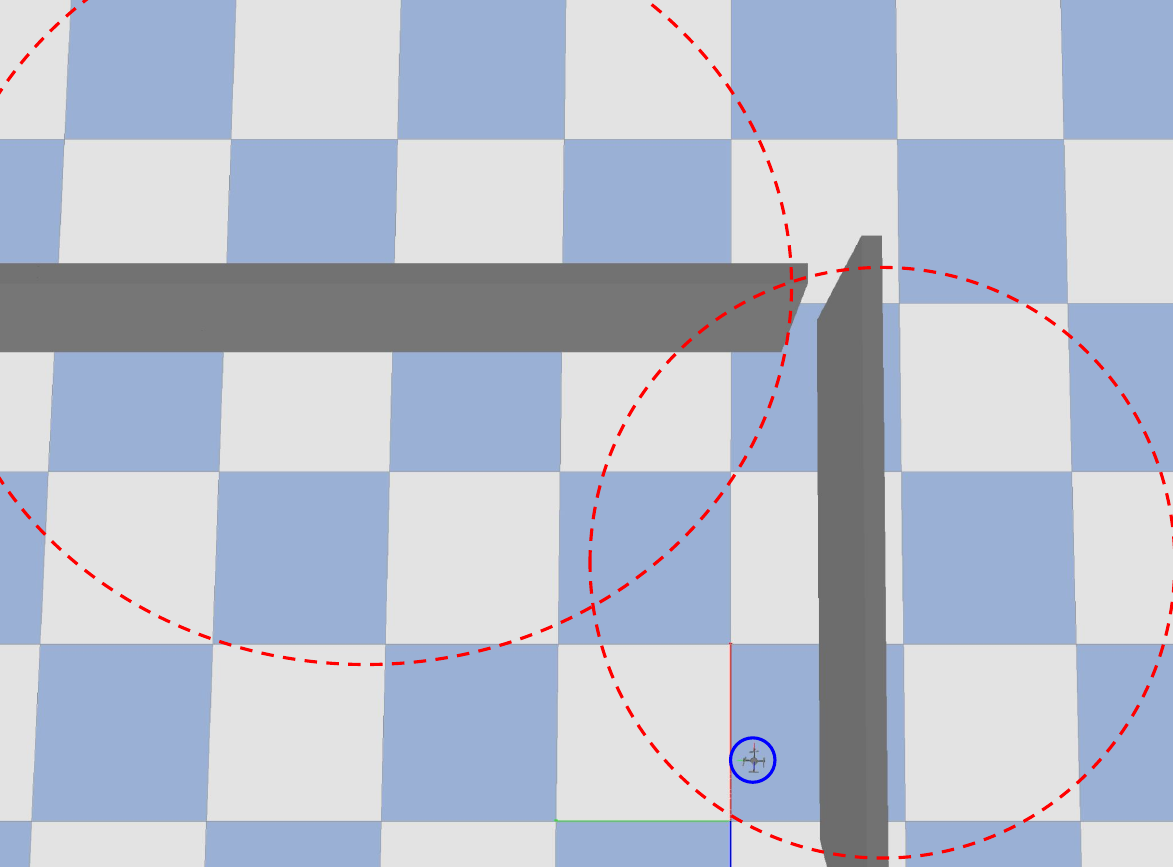}
        \caption{}
        \end{subfigure}
        \begin{subfigure}{\linewidth}
        \includegraphics[width=\linewidth]{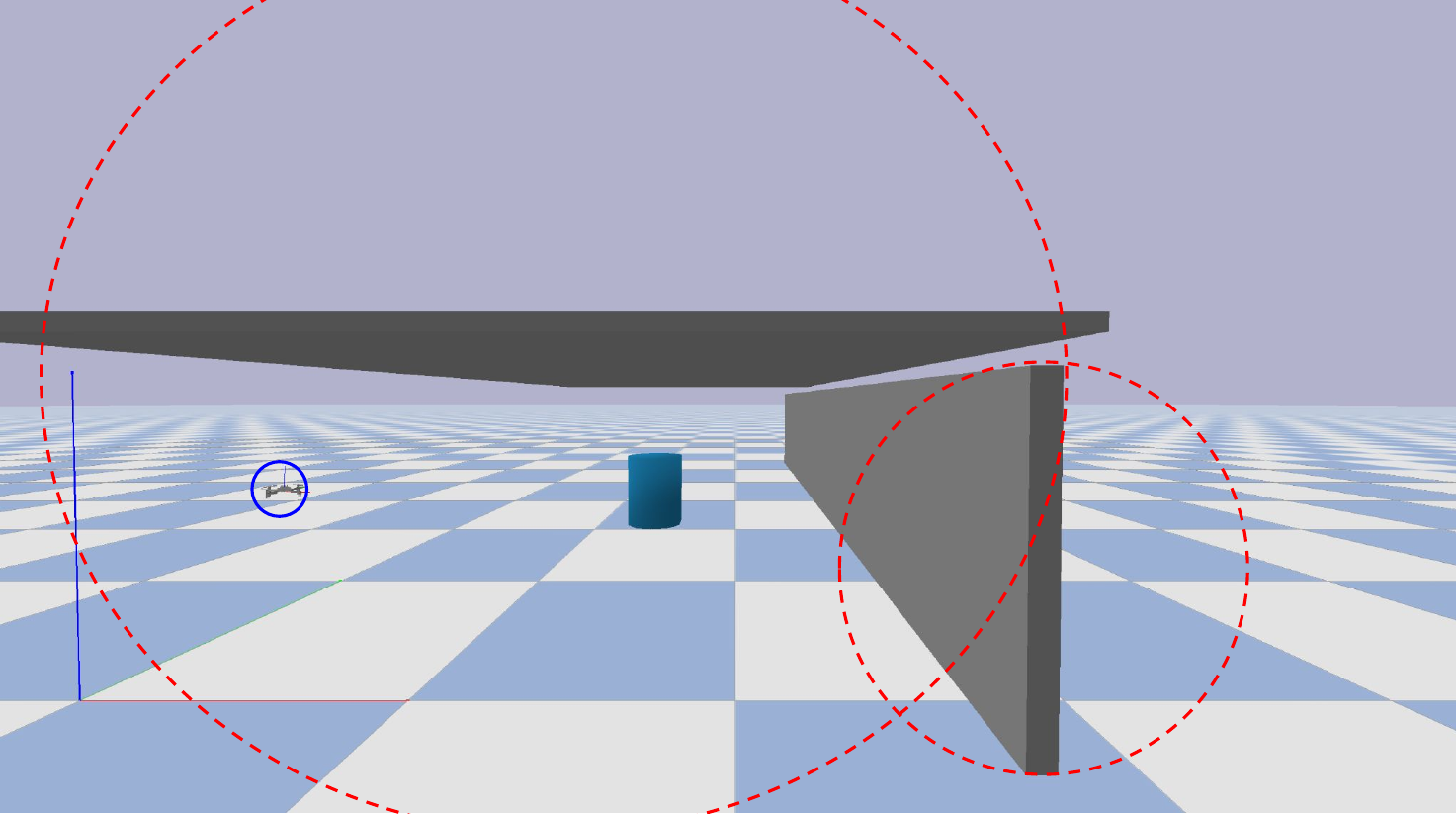}
        \caption{}
        \end{subfigure}
        \end{minipage} 
        \end{center}
        
\caption{Navigation in a cluttered environment using PolyC2BF in case of Quadrotor in 2 different arrangements}
\label{fig:cluttered_walk_quadrotor}
\end{figure}

For both the cases, i.e. quadruped and drone, we have considered the reference control inputs as a simple PD controller for simulating which can be replaced with any state-of-the-art path planning/ trajectory tracking controller.
Constant target velocities were chosen to verify the PolyC2BF-QP. Here, we select the class-$\mathcal{K}$ function $\kappa(h) = \gamma h$ as the identity map i.e. $\gamma=1$. In all the experiments, the position and the velocity of the obstacles are assumed to be measured and always available. Fig. \ref{fig:cluttered_walk_quadruped} and Fig. \ref{fig:cluttered_walk_quadrotor} (a) and (b) shows the quadruped and drone moving in a cluttered environment in front of a very long wall, which would not have been possible with collision cone control barrier function. The red dotted circles in the figures represent the virtual obstacles considered by the collision cone control barrier function (C3BF); thus, employing the C3BF implies that the ego-vehicle's initial position lies within these virtual obstacles, rendering it ineffective in such scenarios.
All the simulations experiments can be viewed in the attached webpage\footnote{\label{note: Videos link}.\url{https://tayalmanan28.github.io/PolyC2BF/}}

\section{Conclusions}
\label{section: Conclusions}
We presented a novel CBF formulation for collision avoidance in cluttered environments and tight spaces. Unlike existing works, the proposed Polygonal Collision Cone based QP formulation (PolyC3BF-QP) allows the vehicle to safely maneuver cluttered envioronments and tight spaces, in real time without any complexity of optimization. This is demonstrated in PyBullet simulations of Quadruped and Quadrotor. Future work includes considering more complex 3D shapes and cases for quadrotors and hardware implementation.

\label{section: References}
\bibliographystyle{IEEEtran}
\bibliography{references.bib}

\end{document}